\documentclass[runningheads]{llncs}
\usepackage[T1]{fontenc}
\usepackage{graphicx}
\usepackage{amsmath}
\usepackage{booktabs}
\usepackage{multirow}
\usepackage{url}

\usepackage{xcolor}
\newcommand{\rev}[1]{{#1}}

\begin{document}
\title{A Black-Box Adversarial Attack on Human Pose Estimation and Keypoint-Based Action Recognition Models}
\titlerunning{A Black-Box Adversarial Attack on Human Pose Estimation}

\author{Kacper Mroczek\inst{1} \and
Michal Kepski\inst{1}\orcidID{0000-0003-1225-9143}}
\authorrunning{K. Mroczek and M. Kepski}
\institute{University of Rzeszów, 35-959 Rzeszów, Poland \\
\email{mkepski@ur.edu.pl}}
\maketitle              
\begin{abstract}

Human pose estimation and keypoint-based action recognition models are increasingly deployed as components of video understanding pipelines, yet their vulnerability to adversarial attacks remains insufficiently studied.
Temporally coherent black-box attacks have been previously studied in visual object tracking, where the attack feedback can be defined using bounding-box overlap measures such as Intersection over Union (IoU). However, human pose estimation produces keypoint configurations rather than enclosing boxes, making box-level similarity poorly suited for measuring pose degradation. We propose OKS Attack, a decision-based black-box attack that uses Object Keypoint Similarity (OKS) as the attack feedback signal, directly targeting the spatial structure of human poses rather than their enclosing boxes.

Experiments on the Penn Action dataset show that OKS Attack \rev{consistently reduces pose quality across evaluated pose estimators, with mean OKS decreases ranging from 0.0802 to 0.1494. In a downstream cross-dataset action-recognition evaluation, the attack reduces accuracy by 6.18 to 13.86 percentage points and outperforms query-matched random-noise perturbations. The attack is effective across both top-down and single-stage pose estimation models. The source code will be made publicly available at \url{https://github.com/KacperM33/OKS_attack}.}

\end{abstract}

\section{Introduction}

Recently, deep neural networks (DNNs) have significantly advanced visual understanding in many tasks such as image classification, object
detection, semantic segmentation, etc. They have also been successfully applied to video analysis in tasks like object tracking or action recognition. 

\begin{figure}
\includegraphics[width=\textwidth]{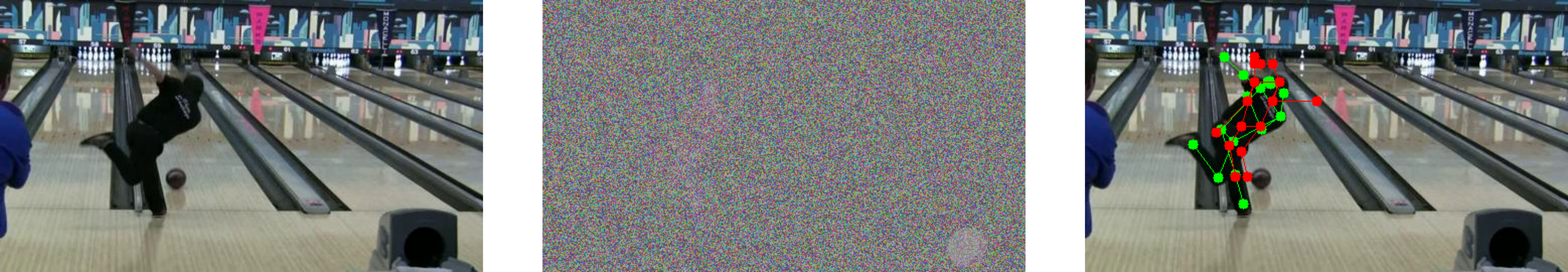}
\caption{Overview of the proposed attack. The original frame (left) is perturbed with adversarial noise (middle) optimized using keypoint-similarity-based feedback. The \rev{limited} perturbation shifts the predicted pose away from the clean estimate, shown by the clean (green) and adversarial (red) skeletons.}
 \label{fig1}
\end{figure}

Simultaneously, it has been shown that deep neural networks are vulnerable to adversarial examples \cite{szegedy_intriguing_2013} which, by the presence of carefully selected noise, can degrade DNN performance leading to incorrect predictions. Since then, a great deal of research has been performed in this area showing that vulnerability to adversarial samples exists in a wide range of computer vision settings \cite{croce2021robustbench,eykholt_robust_2018,wei_towards_2022,xie_adversarial_2017}. The majority of the research focused on image understanding, but video-based tasks, mainly object tracking, received some attention \cite{jia_iou_2021}.

Human pose estimation (HPE) is a fundamental computer vision task that impacts applications such as action recognition, person reidentification, and video anomaly detection. Adversarial attacks on HPE can propagate errors to downstream tasks, reducing their performance. Although image classification has seen extensive work on adversarial attacks, only a few works consider an adversary goal to mislead an HPE model. Most focus on single images, not videos, and do not examine the influence of adversarial samples on downstream tasks.

We propose a decision-based black-box adversarial attack that targets 2D HPE models. We adapt the temporally coherent method proposed in \cite{jia_iou_2021} from box overlap to keypoint similarity \rev{and reformulate its feedback mechanism for structured human-pose outputs}. For each frame, the attack compares two pose predictions: one from the clean input frame and one from its perturbed counterpart. The Object Keypoint Similarity between these predictions serves as the optimization feedback. At each iteration, we sample a set of candidate tangential perturbations with equal noise magnitude and evaluate their effect on the predicted pose. The candidate that yields the lowest OKS score is selected as the most effective perturbation direction. We then slightly extend this perturbation along the normal direction and combine both components to obtain the perturbation for the current iteration. The perturbation is propagated to subsequent frames as initialization to reinforce the temporal aspect of the attack. 

We evaluate OKS Attack on diverse 2D HPE models, including top-down heatmap-based pipelines, as well as single-stage estimators such as YOLO-Pose \cite{maji_yolo-pose_2022}. In addition, we study how adversarially perturbed poses affect downstream keypoint-based action recognition. The main contributions of this work are summarized as follows:
\begin{itemize}
\item \rev{We propose OKS Attack, a task-specific adaptation of decision-based black-box attacks to 2D HPE, where candidate perturbations are guided by scale-normalized Object Keypoint Similarity rather than bounding-box overlap.}
\item We study the impact of adversarially perturbed poses on keypoint-based action recognition.
\item We evaluate the attack across top-down heatmap-based and single-stage pose estimators \rev{and compare it with query-matched random-noise perturbations}.
\item We conduct an ablation study separating the attack's effect on the person detector and the keypoint estimator in top-down pipelines.
\end{itemize}

\section{Related Work}

\subsection{Human Pose Estimation}

Human pose estimation aims to infer the spatial configuration of human body parts from visual input. Since the breakthrough of deep learning-based HPE, introduced by DeepPose \cite{toshev_deeppose_2014}, neural network approaches have been extensively studied in the computer vision literature. HPE is commonly divided into 2D and 3D estimation. 3D HPE aims to predict body joint locations in 3D space, enabling applications such as animation, virtual reality, and sports analysis \cite{zheng_deep_2023}. 

2D HPE estimates the position of body keypoints in the image plane. Existing methods are mainly categorized into top-down and bottom-up approaches \cite{zheng_deep_2023}. Top-down methods use a two-stage pipeline: they first employ a person detector to obtain a set of boxes and perform person pose estimation for each detection. These estimators are typically regression-based \cite{Li_2021_ICCV,toshev_deeppose_2014} or heatmap-based \cite{Sun_2019_CVPR,Yang_2021_ICCV}. Bottom-up methods detect candidate joints and group them into individual poses using part-association strategies \cite{Cheng_2020_CVPR,OpenPose_2021}. Recent single-stage alternatives directly predict person instances and keypoints in one forward pass \cite{maji_yolo-pose_2022,Lu_2024_CVPR}.

\subsection{Keypoint-Based Action Recognition}

Action recognition assigns action labels to people in images or videos. Methods are commonly grouped by input representation. RGB-based approaches operate directly on video frames and learn spatio-temporal appearance and motion features, while pose-based methods use human keypoints from estimated 2D poses or 3D skeleton data.
In 2D pose-based action recognition, methods differ mainly in how keypoints are represented over time. Some approaches model sequences of 2D joint coordinates directly, using temporal or attention-based architectures to capture motion patterns \cite{mazzia_action_2022}. Other works convert 2D keypoints into spatio-temporal heatmap volumes and apply 3D CNNs, as in PoseC3D \cite{duan_revisiting_2022}. Skeleton-based action recognition commonly relies on graph-based models due to the natural joint-bone structure \cite{yan_spatial_2018}.

\subsection{Adversarial Attacks}

Adversarial examples were first studied in image classification, where small perturbations can change the prediction of deep neural networks \cite{szegedy_intriguing_2013}. Subsequent work showed that adversarial vulnerability is not limited to CNN-based classifiers, but also affects other architectures and vision tasks, including vision transformers \cite{wei_towards_2022}. Adversarial attacks are commonly divided into digital attacks, which add subtle perturbations directly to input data, and physical attacks, which introduce adversarial patterns through real-world objects such as patches or stickers \cite{wei_physical_2024}. Depending on the assumed access to the target model, attacks are typically categorized as white-box or black-box. In the latter setting, the adversary does not use gradients and relies only on the model output.  

Compared with image-level tasks, adversarial attacks on video understanding remain less explored. In visual object tracking, IoU Attack is particularly relevant, as it uses bounding-box Intersection over Union as feedback in a black-box setting \cite{jia_iou_2021}. However, such feedback is naturally tied to box-based tracking outputs and does not capture errors in structured keypoint configurations. 

Adversarial robustness of human pose estimation has received comparatively limited attention, with most studies focusing on still-image pose estimators rather than downstream video tasks \cite{Jain_2019_CVPR_Workshops}. \rev{Recent HPE-specific attacks include local imperceptible perturbations against pose estimation networks \cite{liu_local_2023} and transferable attacks designed to improve cross-model transferability \cite{chen_transferable_2026}. Other recent work has explored object keypoint similarity in adversarial attacks on human pose estimation, but under a white-box threat model \cite{mu_generating_2026}. In the action-recognition setting, black-box attacks and robustness studies commonly perturb skeleton or joint sequences directly, rather than the visual input before pose extraction \cite{Diao_2021_CVPR}. In contrast, our work attacks video frames before 2D pose estimation and evaluates how the resulting pose perturbations affect downstream keypoint-based action recognition.}

To the best of our knowledge, no prior work studies a temporally coherent decision-based black-box attack that uses OKS as feedback for 2D HPE from video frames and evaluates its impact on downstream keypoint-based action recognition.

\section{Proposed method}

\subsection{Overview}

In a decision-based black-box setup, the adversary has no access to the target model architecture, parameters, gradients, or confidence scores. Instead, the attack can only query the model and observe its final output or decision. Given a benign input $x$, the goal is to construct an adversarial example $x_{\mathrm{adv}} = x + \delta$ by iteratively updating the perturbation $\delta$ based on the observed model responses.

In our work, we adopt the attack framework introduced by \cite{brendel__decision-based_2018} and later adapted by \cite{jia_iou_2021}, which starts from a large adversarial perturbation and iteratively reduces its magnitude while preserving the adversarial effect. In the pose estimation setting we measure the attack effect through the deviation between the pose predicted on the clean frame and the pose predicted on the perturbed frame. The clean prediction serves as a reference pose, which makes the attack applicable without ground-truth annotations at test time. To quantify this deviation, we use Object Keypoint Similarity, a standard keypoint-level similarity measure that accounts for the spatial distance between corresponding joints. A successful perturbation should reduce the OKS between clean and adversarial pose predictions while remaining visually small. 

Since the input is a video sequence, the attack also exploits temporal continuity between consecutive frames. The perturbation found for the current frame is reused to initialize the attack on subsequent frames, encouraging temporally consistent degradation. This framework is model-agnostic and can be applied to different 2D pose estimation paradigms, as long as the target model returns keypoint predictions for each frame.

\subsection{OKS Attack}

Given a clean original image, we first add strong random noise to obtain an initial perturbed image for which the OKS between clean and perturbed pose predictions is low. Moving from the clean image toward this noisy image generally decreases OKS while increasing the perturbation magnitude. OKS Attack therefore searches for a perturbation that sufficiently reduces pose similarity while keeping the added noise as small as possible. 

Let $I_t$ denote the original image at the $t$-th video frame and let $F(\cdot)$ be the target 2D pose estimator. We denote the clean pose prediction by
\begin{equation}
    P_t^{0} = F(I_t),
\end{equation}
where $P_t^{0} = \{p_{t,i}^{0}\}_{i=1}^{M}$ contains $M$ predicted body keypoints. At the $k$-th iteration of the attack, the current perturbed image is denoted by $I_t^{(k)}$. We randomly sample $n$ tangential perturbations $\eta_t^j$, $j \in \{1,\ldots,n\}$, and normalize them so that they preserve the current perturbation magnitude:
\begin{equation}
    D(I_t, I_t^{(k)}) = D(I_t, I_t^{(k)} + \eta_t^j),
\end{equation}
where $D$ is a pixel-wise distance measure between two images. For each candidate image $I_t^{(k)} + \eta_t^j$, we query the pose estimator and obtain a candidate pose
\begin{equation}
    P_t^j = F(I_t^{(k)} + \eta_t^j).
\end{equation}

To evaluate the effect of each candidate perturbation, we use the COCO-style Object Keypoint Similarity (OKS) \cite{Ronchi_2017_ICCV}. Given two poses $P=\{p_i\}_{i=1}^{M}$ and $Q=\{q_i\}_{i=1}^{M}$, we define
\begin{equation}
    \operatorname{OKS}(P,Q) =
    \frac{
    \sum_{i=1}^{M} m_i
    \exp\left(
    - \frac{\|p_i-q_i\|_2^2}{2s^2\kappa_i^2}
    \right)
    }{
    \sum_{i=1}^{M} m_i
    },
\end{equation}

where $m_i$ indicates whether the $i$-th reference keypoint is visible, $s$ is the person scale, and $\kappa_i$ is the keypoint-specific normalization constant used in COCO evaluation. In our setting, the clean prediction $P_t^0$ is used as the reference pose, which allows the attack to operate without ground-truth keypoint annotations.

Following the temporal formulation of IoU Attack, we combine a spatial and temporal OKS score:
\begin{equation}
    S_{\mathrm{OKS}}^j =
    \lambda S_{\mathrm{spatial}}^j +
    (1-\lambda) S_{\mathrm{temporal}}^j,
\end{equation}
where
\begin{equation}
    S_{\mathrm{spatial}}^j =
    \operatorname{OKS}(P_t^{0}, P_t^j),
\end{equation}
and
\begin{equation}
    S_{\mathrm{temporal}}^j =
    \operatorname{OKS}(P_{t-1}^{0}, P_t^j).
\end{equation}
For the first frame, only the spatial term is used. The parameter $\lambda$ balances the influence of current-frame pose degradation and temporal consistency. Since lower OKS indicates a larger deviation from the reference pose, we select the candidate perturbation with the lowest score:
\begin{equation}
    j^{*} = \arg\min_{j \in \{1,\ldots,n\}} S_{\mathrm{OKS}}^j.
\end{equation}

The selected tangential perturbation $\eta_t^{j^{*}}$ is then composed with a small normal perturbation $\nu_t^{j^{*}}$, producing the updated adversarial image
\begin{equation}
    I_t^{(k+1)} =
    \Pi_{\mathcal{I}}
    \left(
    I_t^{(k)} + \eta_t^{j^{*}} + \nu_t^{j^{*}}
    \right),
\end{equation}
where $\Pi_{\mathcal{I}}$ clips the result to the valid image range. After the optimization for frame $t$ is completed, the final perturbation is propagated to the next frames as initialization, encouraging temporally consistent degradation and reducing the number of queries required for subsequent frames.

\section{Experiments}

We validate the performance of our method on four pose estimation networks and one keypoint-based action recognition model, using the challenging Penn Action Dataset. Detailed results are provided as follows.

\subsection{Dataset}

\textbf{Penn Action Dataset} ~\cite{Zhang_2013_ICCV} is a sports action dataset collected by the University of Pennsylvania. It contains 2,326 videos covering 15 action classes, with each frame annotated using 13 human keypoints. Penn Action is particularly suitable for our study, as it provides both frame-level keypoint annotations and action labels, allowing us to evaluate the attack's effect on pose estimation and downstream action recognition within a single benchmark.

\subsection{Experiment setup}

\subsubsection{Human pose estimation models.} In order to validate the generality of our black-box adversarial attack, we choose four representative 2D HPE models with different structures: ResNet-50 following the SimpleBaseline2D top-down heatmap architecture \cite{xiao2018simple}, lightweight MobileNetV2 architecture \cite{sandler2018mobilenetv2} and two variants of single-stage YOLO-Pose \cite{maji_yolo-pose_2022}: YOLO-Pose S and Yolo-Pose M. For pose estimation, we use publicly available checkpoints from MMPose \cite{mmpose2020} library trained on COCO Dataset \cite{lin2014microsoft}.

\subsubsection{Action recognition model.} We use PoseC3D with a Pose-SlowOnly R50 backbone as a downstream 2D pose-based action recognition model. The model converts estimated 2D keypoints into spatio-temporal heatmap volumes and classifies actions using a 3D CNN \cite{duan_revisiting_2022}. Since the model is not trained on Penn Action, we map the predicted class labels to the corresponding Penn Action categories and use it to evaluate whether pose perturbations affect action recognition in a cross-dataset setting.

\subsubsection{Implementation details.} 

We initialize the attack by adding Gaussian noise to the clean frame, which defines a noisy reference direction for the perturbation search.
\rev{We use $K=15$ iterations with $N=30$ candidates per iteration, resulting in a maximum query budget of $2+2KN$ model queries per frame.}
The score combines spatial and temporal OKS terms with weight $\lambda=0.9$, prioritizing the current-frame pose degradation while encouraging temporal consistency. To keep perturbations visually limited, we clip them pixel-wise to \rev{10.0} in the centered image space and clip final adversarial frames to the valid image range. Across consecutive frames, the previous perturbation is reused with weight 0.9 and smoothed with decay factor 0.8.
Since Penn Action annotates a single target person, while some frames contain multiple people, we use the ground-truth bounding box to select the relevant prediction. For top-down models, only the detected box with the highest IoU to the ground-truth box is passed to the pose estimator; for single-stage models, we select the predicted skeleton whose enclosing box best overlaps the ground-truth person box. Before computing OKS, we discard keypoints not present in the 13-keypoint Penn Action format, while action recognition uses the full predicted keypoint set. 
Due to the computational cost and the number of experimental variants, we evaluate all methods on a randomly selected stratified \rev{35\%} subset of the Penn Action test set.

\subsection{Overall Results}

\subsubsection{Pose estimation results.} 

We first evaluate the direct impact of the attack on 2D HPE, independently of action recognition. For each model, we compare clean and adversarial pose predictions using  the average OKS and the average number of lost person detections. Lower OKS indicates a larger deviation from the reference pose, while more losses indicate missing or unreliable pose predictions. \rev{Since HPE produces continuous keypoint outputs rather than discrete class labels, we treat pose-level attack effectiveness as a continuous degradation measured by OKS decrease, rather than as a binary success/failure outcome}. We report results for all videos and separately for samples with unchanged or changed downstream action predictions.

\begin{table}[t]
\centering
\rev{
\caption{Mean OKS for clean and adversarial pose predictions. Results are reported for all videos, action-changed videos, and action-unchanged videos. Random noise denotes a query-matched baseline with the same perturbation budget as OKS Attack.}
\label{tab:oks_hpe_results}
\setlength{\tabcolsep}{2.25pt}
\small
\begin{tabular}{lccccccccc}
\toprule
\multirow{3}{*}{2D HPE} 
& \multicolumn{9}{c}{Mean Object Keypoint Similarity} \\
\cmidrule(lr){2-10}
& \multicolumn{3}{c}{All videos}
& \multicolumn{3}{c}{Action changed}
& \multicolumn{3}{c}{Action unchanged} \\
\cmidrule(lr){2-4}
\cmidrule(lr){5-7}
\cmidrule(lr){8-10}
& Orig & \shortstack{OKS\\Attack} & \shortstack{Rand.\\noise}
& Orig & \shortstack{OKS\\Attack} & \shortstack{Rand.\\noise}
& Orig & \shortstack{OKS\\Attack} & \shortstack{Rand.\\noise} \\
\midrule
ResNet-50
& 0.7361 & \textbf{0.6045} & 0.6924
& 0.6189 & \textbf{0.4200} & 0.5078
& 0.7633 & \textbf{0.6472} & 0.7154 \\
MobileNetV2
& 0.6931 & \textbf{0.5437} & 0.6635
& 0.4674 & \textbf{0.3444} & 0.4128
& 0.7558 & \textbf{0.6221} & 0.6997 \\
YOLO-Pose M
& 0.7547 & \textbf{0.6745} & 0.7121
& 0.6592 & \textbf{0.5362} & 0.5911
& 0.7857 & \textbf{0.7194} & 0.7418 \\
YOLO-Pose S
& 0.7262 & \textbf{0.6111} & 0.6623
& 0.6355 & \textbf{0.4785} & 0.5303
& 0.7812 & \textbf{0.6920} & 0.7123 \\
\bottomrule
\end{tabular}
}
\end{table}

Table~\ref{tab:oks_hpe_results} shows that OKS Attack consistently reduces pose estimation quality for all evaluated models. Across all videos, the largest OKS decrease is observed for MobileNetV2, where the mean OKS drops from \rev{0.6931} to \rev{0.5437}, followed by ResNet-50 and YOLO-Pose S. The degradation is substantially stronger for samples in which the action prediction changes. This suggests that successful action-prediction changes are associated with larger pose-estimation errors. We also observe that samples leading to action changes tend to have lower clean OKS even before the attack. This indicates that sequences for which the pose estimator is already less stable are more vulnerable to adversarial perturbations.

\subsubsection{Action recognition results.} 

We next evaluate whether pose perturbations generated by OKS Attack affect downstream keypoint-based action recognition. \rev{Table~\ref{tab:action_and_distortion} reports action-recognition accuracy for clean pose sequences, OKS Attack, and query-matched random-noise perturbations.}

\begin{table}[t]
\centering
\rev{
\caption{Action-recognition accuracy and distortion metrics for OKS Attack.}
\label{tab:action_and_distortion}
\small
}
\begin{minipage}[t]{0.57\textwidth}
\centering
\rev{
\begin{tabular}{lccc}
\toprule
2D HPE model & Clean acc. & OKS Attack & Random noise \\
\midrule
ResNet-50    & 79.03\% & 72.85\% & 78.06\% \\
MobileNetV2  & 73.52\% & 66.30\% & 72.38\% \\
YOLO-Pose M  & 76.80\% & 69.87\% & 72.53\% \\
YOLO-Pose S  & 71.73\% & 57.87\% & 65.33\% \\
\bottomrule
\end{tabular}
}
\end{minipage}
\hfill
\begin{minipage}[t]{0.36\textwidth}
\centering
\rev{
\begin{tabular}{lcc}
\toprule
2D HPE model & PSNR & SSIM \\
\midrule
ResNet-50    & 29.83 & 0.69 \\
MobileNetV2  & 30.53 & 0.71 \\
YOLO-Pose M  & 29.32 & 0.68 \\
YOLO-Pose S  & 29.46 & 0.68 \\
\bottomrule
\end{tabular}
}
\end{minipage}

\end{table}

\begin{figure}
\includegraphics[width=\textwidth]{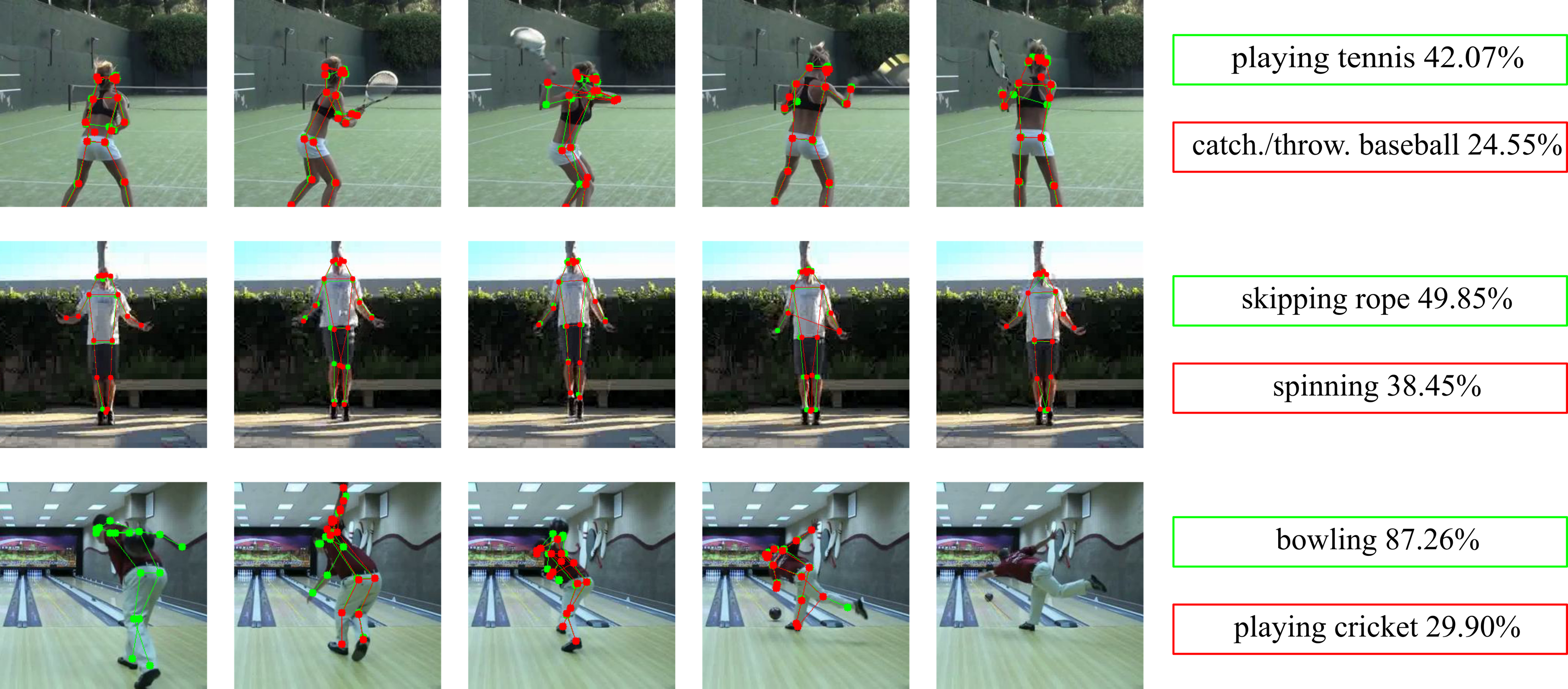}
\caption{Examples of action prediction changes. Each row shows frames sampled from one video at a fixed interval of $\Delta=10$ frames. Clean and adversarial pose predictions are overlaid in green and red; the absence of a skeleton indicates that no pose was detected. Boxes report clean and adversarial predictions with confidence scores.}
\label{fig:action_examples}
\end{figure}

\rev{OKS Attack reduces downstream accuracy for all evaluated pose estimators, with drops ranging from 6.18 percentage points for ResNet-50 to 13.86 for YOLO-Pose S. For the evaluated single-stage models, it also causes stronger degradation than query-matched random noise, indicating that the effect is not due only to the perturbation budget.} Since the downstream action-recognition model is used off-the-shelf and is not trained on Penn Action, these results should be interpreted as a cross-dataset robustness evaluation of a pretrained pose-based recognition pipeline rather than as a supervised Penn Action benchmark. Figure~\ref{fig:action_examples} shows selected affected sequences, where small skeleton shifts are sufficient to change the downstream action prediction.

\subsubsection{Distortion metrics.} 

Table~\ref{tab:action_and_distortion} also reports PSNR and SSIM between clean and adversarial frames for OKS Attack. Across the evaluated pose estimators, PSNR ranges from 29.32 to 30.53 and SSIM from 0.68 to 0.71. These values show that the perturbations are bounded and visually limited, while still introducing measurable image-level distortion.

\subsection{Ablation Study}

For top-down pose estimation pipelines, the pose quality depends on two components: the person detector and the single-person pose estimator. We observed that OKS Attack also affects the detection stage: the average number of lost person detections per sequence increased from \rev{8.63} to \rev{21.6} for MobileNetV2 and from \rev{4.91} to \rev{16.32} for ResNet-50. This suggests that the observed degradation of HPE may result from both inaccurate person localization and errors introduced by the pose estimator itself.

To disentangle these effects, we conduct an ablation study on top-down methods using two additional evaluation settings. In the \textit{NoDet-Clean} setting, the detector is not applied to adversarially perturbed frames; instead, we use the bounding boxes obtained by running the detector on the corresponding clean frames. This removes the direct influence of adversarial noise on the detector while preserving realistic detector outputs. In the \textit{NoDet-GT} setting, the detector is removed entirely and the pose estimator receives ground-truth person bounding boxes from the dataset. This setting isolates the robustness of the keypoint estimator itself, independent of detection errors.

The ablation results in Table~\ref{tab:ablation_detector} show that the attack affects both stages of the top-down pipeline. For ResNet-50, the full pipeline achieves an OKS decrease of \rev{0.1316}, while using clean-frame detector boxes reduces the drop to 0.0535, and replacing the detector with ground-truth boxes further reduces it to 0.0411. MobileNetV2 follows the same pattern, but the difference between the detector-controlled settings is smaller. These results indicate that the degradation partly comes from attacking the person detector, while the remaining OKS drop in detector-controlled settings shows that the pose estimator is also affected. The non-zero OKS decrease for single-stage YOLO-Pose models supports this conclusion, as they do not use a separate top-down detector.

\begin{table}[t]
\centering
\caption{Ablation study on the source of pose degradation. For top-down models, we compare the full pipeline with detector-controlled variants. YOLO-Pose models are included as single-stage references.}
\label{tab:ablation_detector}
\setlength{\tabcolsep}{4pt}
\small
\begin{tabular}{lllc}
\toprule
Model & Setting & Box source & OKS decrease \\
\midrule
ResNet-50 & Full pipeline & Attacked detector & \rev{0.1316} \\
ResNet-50 & NoDet-Clean & Clean-frame detector & \rev{0.0535} \\
ResNet-50 & NoDet-GT & Ground-truth box & \rev{0.0411} \\
\midrule
MobileNetV2 & Full pipeline & Attacked detector & \rev{0.1494} \\
MobileNetV2 & NoDet-Clean & Clean-frame detector & \rev{0.0852} \\
MobileNetV2 & NoDet-GT & Ground-truth box & \rev{0.0712} \\
\midrule
YOLO-Pose S & Single-stage & Not applicable & \rev{0.1151} \\
YOLO-Pose M & Single-stage & Not applicable & \rev{0.0802} \\
\bottomrule
\end{tabular}
\end{table}

\section{Conclusions}

In this paper, we introduced OKS Attack, a decision-based black-box adversarial attack for 2D human pose estimation in video. Instead of relying on bounding-box overlap, the proposed method uses Object Keypoint Similarity as a task-specific feedback, which makes the attack directly aligned with the structured keypoint output of pose estimators. The attack is model-agnostic and can be applied to both top-down and single-stage pose estimators.

Experiments on Penn Action show that OKS Attack consistently degrades pose estimation quality, with mean OKS decreases ranging from 0.0802 to 0.1494 across evaluated models. The comparison with query-matched random-noise perturbations indicates that the observed degradation is not explained by the perturbation budget alone. The ablation study shows that, in top-down pipelines, the attack affects both the person detector and the keypoint estimator, while results on single-stage YOLO-Pose models confirm that the attack is not limited to detector degradation. Future work will investigate stronger temporal attack objectives, physical-world perturbations, and robustness evaluation with action recognition models trained directly on the target dataset.

\subsubsection{Acknowledgements} We gratefully acknowledge Polish high-performance computing infrastructure PLGrid (HPC Center: ACK Cyfronet AGH) for providing computer facilities and support within computational grant no. PLG/2026/019605.

%
%
%
\bibliographystyle{splncs04}
\bibliography{references}

\end{document}